\documentclass[sigconf]{acmart}

\usepackage{booktabs} 
\usepackage{subcaption}
\usepackage{graphicx}
\usepackage{hyperref}
\usepackage{multirow}
\usepackage{multicol}
\usepackage[export]{adjustbox}
\usepackage{placeins}
\usepackage{breqn}

\definecolor{red_plot}{HTML}{D62728}
\definecolor{green_plot}{HTML}{2CA02C}

\copyrightyear{2018} 
\acmYear{2018} 
\setcopyright{acmlicensed}
\acmConference[GECCO '18]{Genetic and Evolutionary Computation Conference}{July 15--19, 2018}{Kyoto, Japan}
\acmBooktitle{GECCO '18: Genetic and Evolutionary Computation Conference, July 15--19, 2018, Kyoto, Japan}
\acmPrice{15.00}
\acmDOI{10.1145/3205455.3205567}
\acmISBN{978-1-4503-5618-3/18/07}

\begin{document}
\title{Real-World Evolution Adapts Robot Morphology and Control to Hardware Limitations}

\author{T\o nnes F. Nygaard}
\affiliation{
  \institution{University of Oslo, Norway}
}
\email{tonnesfn@ifi.uio.no}

\author{Charles P. Martin}
\affiliation{
  \institution{University of Oslo, Norway}
}
\email{charlepm@ifi.uio.no}

\author{Eivind Samuelsen}
\affiliation{
  \institution{University of Oslo, Norway}
}
\email{eivinsam@ifi.uio.no}

\author{Jim Torresen}
\affiliation{
  \institution{University of Oslo, Norway}
}
\email{jimtoer@ifi.uio.no}

\author{Kyrre Glette} 
\affiliation{
  \institution{University of Oslo, Norway}
}
\email{kyrrehg@ifi.uio.no}

\renewcommand{\shortauthors}{T.F.Nygaard et al.}

\begin{abstract}
For robots to handle the numerous factors that can affect them in the real world, they must adapt to changes and unexpected events.
Evolutionary robotics tries to solve some of these issues by automatically optimizing a robot for a specific environment.
Most of the research in this field, however, uses simplified representations of the robotic system in software simulations.
The large gap between performance in simulation and the real world makes it challenging to transfer the resulting robots to the real world.
In  this paper, we apply real world multi-objective evolutionary optimization to optimize both control and morphology of a four-legged mammal-inspired robot.
We change the supply voltage of the system, reducing the available torque and speed of all joints, and study how this affects both the fitness, as well as the morphology and control of the solutions.
In addition to demonstrating that this real-world evolutionary scheme for morphology and control is indeed feasible with relatively few evaluations,
we show that evolution under the different hardware limitations results in comparable performance for low and moderate speeds, and that the search achieves this by adapting both the control and the morphology of the robot.

\end{abstract}

%
%



\maketitle

\vspace{-2mm}
\vspace{2mm}

\begin{figure}
\vspace{6mm}
\centering
  \begin{subfigure}{0.23\textwidth}
    \includegraphics[width=1.0\textwidth]{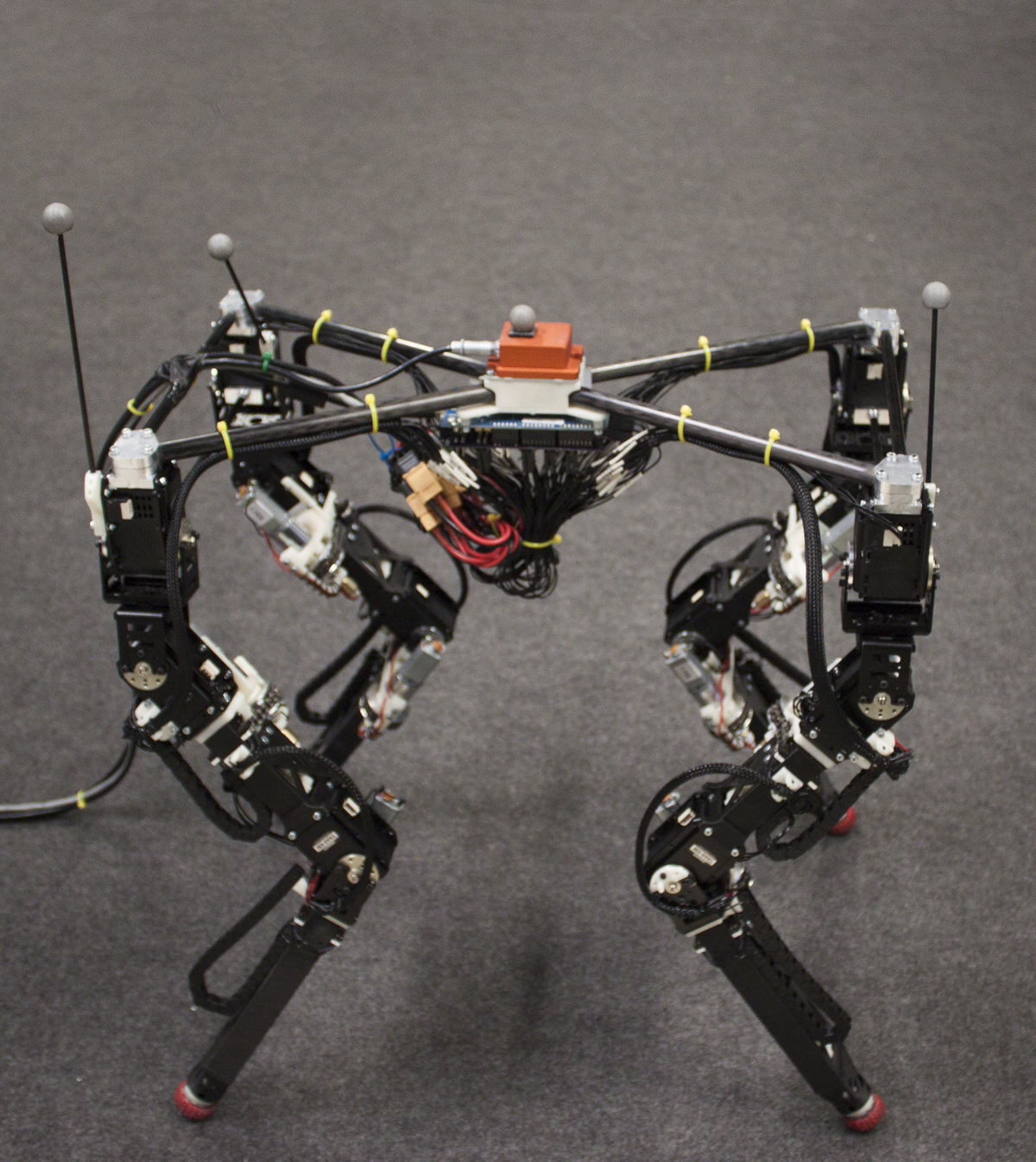}
    \caption{Shortest possible legs}
  \end{subfigure}
  \begin{subfigure}{0.23\textwidth}
    \includegraphics[width=\textwidth]{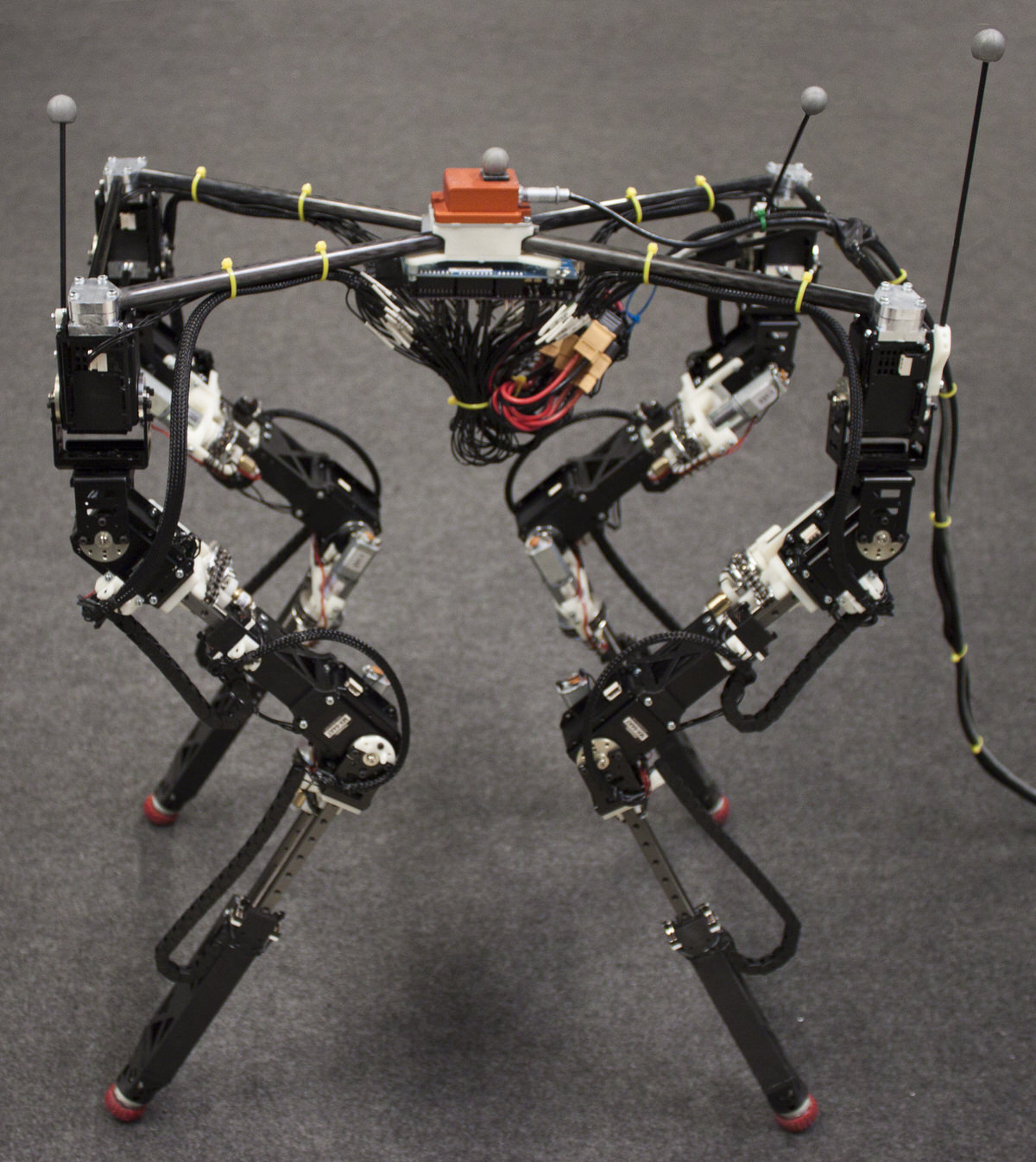}
    \caption{Longest possible legs}
  \end{subfigure}
    \caption{The robot used in this research features self-modifying legs. The length of the two lower limbs of all four legs can be set individually with sub-millimetre accuracy.}
  \label{fig.robot}
  \vspace{-5mm}
\end{figure}

\section{Introduction}
Evolutionary robotics (ER) uses techniques from evolutionary compution to optimize robot control and morphology, and aims to produce robots that are both robust and adaptive \cite{whatwhywhere}. 
One of the biggest challenges in ER, is making the leap from software simulations to experiments evolving real physical robots \cite{grandchallenges}.
Most ER research is done exclusively in simplified physics simulators \cite{mouret201720}.
Projects that transfer evolutionary results to physical robots often face discrepancies in performance between the simulator and the real world, referred to as the \textit{reality gap}.
Evolving in hardware on a real robot bypasses the problem of the reality gap completely, and can even be used for on-line adaptation of the system in its intended environment \cite{horizons}.
Many researchers do not use ER with the intent of producing an optimal robot controller or morphology, but to investigate evolutionary processes.
Real world evolution might, for this objective, yield more realistic results since it exhibits the same noise and unpredictability that other physical systems in nature present.
Evolving in hardware also lets us more closely investigate the embodied cognition aspect of robotics, namely how the interactions between mind, body, and environment affect how a robot solves a task.
One of the biggest challenges in evolving in hardware today, is the long evaluation time required.
This will be reduced with quicker and more accurate evaluation methods, and new production techniques allowing more systems to be run cheaply and efficiently in parallel might offset much of the difference between simulation and real world evaluation we see today. 

In this paper, we investigate the extent to which control and morphology can be adapted by a real-world evolutionary system if the physical conditions of the system change. 
To achieve this, we use a four-legged robot with high-level control and self-reconfigurable morphology in the form of legs with motorised length adjustment, shown in Figure~\ref{fig.robot}.
In our investigations, we evolve the control and morphology of the robot at two different supply voltages, and compare the resulting individuals.
Introducing a change in hardware conditions by turning down the supply voltage reduces both the available speed and torque of all joints by about 20\%.
A reduction in supply voltage would happen naturally to robots with motors directly powered by a depleting battery.
Lower torque or speed can also be caused by internal effects like the temperature of the DC motors or wear and tear on the servo gears, or by external effects such as friction or texture of the walking surfaces, or the weight of the robot's payload.
We also evaluate individuals resulting from the optimal voltage evolutionary run at the reduced voltage, to investigate the reduction in performance and need for adaptation to this limitation.

The results show that lowering the supply voltage of the robot--when it was evolved for the optimal voltage--can significantly impact the performance, with a reduction of 38\% and 17\% to speed and stability respectively.
However, under evolutionary optimization at the reduced voltage, the robot is able to achieve comparable performance at low and moderate speeds to the optimal voltage individuals.
We observe significant changes in both control and morphology between the two groups of individuals to achieve this. 

The contribution of this paper is twofold: First, we demonstrate that evolution finds different morphology and control combinations suitable for our different hardware limitations, entirely by real-world evolution on a robot with self-reconfigurable morphology.
Secondly, we demonstrate that by having a stable platform with high-level control, it is possible to do exploratory multi-objective morphology and control evolution in relatively few evaluations entirely in hardware.
This allows us to investigate complex real-world dynamics not seen in ER experiments relying solely on software simulations.

\section{Background}
This section reviews related work in the area of evolutionary robotics, with a focus on real world evolution and evolution of morphology. 

\subsection{Evolutionary robotics}
Modern specialized robots can be hard to develop, and are often designed by a team of engineers at considerable expense.
Alternatively, in ER, robot controllers and morphologies can be designed or optimized automatically using evolutionary algorithms to identify new solutions \cite{nolfi2002synthesis}.
In general, evolutionary design has been used to optimize a robot's control or morphology in an off-line fashion, before production, and in a different environment than where the robot would be working \cite{horizons}.
The method of embodied evolution uses on-line evolution of robots in the environment where they will be deployed, and thus the robots will be able to react to changes in that environment as they work \cite{watson2002embodied}.
Embodied evolution has, however, almost exclusively been applied to the control of a robot, as very few robots are able to modify their own morphology during an experiment without considerable human intervention.

\subsection{Real world evolution}
Most ER experiments are not performed on physical robots, but on virtual representations in a simplified physics simulator \cite{mouret201720}.
Here, the number and speed of evaluations is only limited by the access to computational power, and thus such experiments can be performed much faster than their real-world equivalent. 
Not only are real-world experiments more expensive in terms of building and maintaining a robot, but there are challenges due to noise in measurements caused by the body of the robot, its dynamic environment, and the interactions between them \cite{silva2016open}.
These advantages make it easy to see the appeal of only using simulations for evolutionary robotics.

One of the biggest challenges with using simulation in evolutionary robots, however, is the reality gap - the discrepancy between measurements of performance in simulation and the real world \cite{4218893}. 
Modern physics simulators have different trade-offs between speed and accuracy, and game-based physics engines often sacrifice accuracy for additional stability \cite{erez2015simulation}.
Even simulators not focused on efficiency or stability can exhibit accuracy that is too low to allow direct transfers of results to real world counterparts.
There are multiple approaches to deal with the reality gap, including adding noise in simulation \cite{jakobi1995noise}, doing most of the evolutionary search in software before doing the last part in hardware \cite{Nolfi:63867}, or by making a model of the disparity between simulation and reality, and use this to guide the search \cite{koos2013transferability}.
Some of these techniques reduce the reality gap significantly, but the difference still makes it challenging to transfer results to the real world - especially as robots are used in more complex environments.
Other techniques guide the search towards individuals in simulation with behaviors that perform closer to their real world counterparts, and this might successfully circumvent some of reality gap problem to the degree where a subset of solutions might be transfered directly to the real world \cite{mouret201720}.
This does, however, limit the results to the small subset of solutions that has accurate performance in the simulator, and the search might therefore be drawn towards simple behaviors without dynamic effects, that are easier to simulate.

Evolving in hardware bypasses the problem of the reality gap completely, and if evolution is performed on the unrestricted system in the environment where it will be serving, also bypasses the problem sometimes seen in simplified or limited experiments in hardware as well \cite{gongora2009robustness}.
Evolution in hardware is most often done off-line to perform a one-time adaptation to a new task or environment, but can also been done constantly in an on-line fashion to continuously adapt to both slow and abrupt changes to the robot itself or its environment \cite{horizons}.
There are several sources in the real world that contribute to uncertainty and noise in measurements of performance, but these are in many cases connected to the, often very complex, interactions between the control, body, and environment.
Being able to study the synergy between these and see how a robot is able to exploit them separately and together to solve a given task is not possible in a simplified physics simulator.

\subsection{Evolution of robot morphology} 
Evolutionary robotics can be used to evolve morphology and adapt a robot's body to the task it is solving, and the environment where it is doing it.
It can even make the evolution of control quicker, and result in more robust gaits \cite{Bongard1234}.
The field of artificial life evolves virtual creatures, closely related to evolution of robot bodies, but is mainly concerned with the study of the biological processes behind the evolution, and experiments are not done with the intention of producing hardware versions of the resulting bodies \cite{Lehman:2011:EDV:2001576.2001606}.
Most work in evolving virtual creatures is done in simulation alone, one of the earliest examples being Sims' work evolving bodies represented by three dimensional boxes \cite{sims1994evolving}.
This has also been done in later work \cite{4218893}, and expanded to more advanced creatures \cite{komosinski2000world}, though there have been several challenges related to the scalability of these techniques \cite{cheney2016difficulty}.
Evolution of morphology in robotics has also mostly been done in simulation, though the models used are more realistic than the virtual creature counterparts, and the intention is most often to end up with results that could be transferred to the real world.
There are many examples of work evolving the morphology of different types of robots, for instance wheeled robots \cite{542394}, legged robots \cite{tonnesfn_evostar17}, or even soft robots \cite{rieffel2014growing}.
Morphology can also be evolved in modular robotics \cite{zykov2007evolved}, though this most often refers to changing the way static modules are assembled.

There are some examples of evolution of robot morphology in simulation, where a select few morphologies are transferred for testing in the real world, including both legged \cite{samuelsen2013hox} and more non-traditional designs \cite{933266}, but these require excessive human intervention for each morphology tested in the real world.
There are examples of morphological evolution in hardware directly as well, but many require excessive human intervention to build and assemble new morphologies \cite{milan17}, use slow external reconfiguration of modular systems \cite{vujovic2017evolutionary}, or no mechanical reconfiguration at all \cite{827631}.
There have been examples of real-world robot evolution with self-modifying morphology, but only using the dynamic body to speed up or improve the evolution of controllers for a single optimal body \cite{Bongard1234}.
The authors are not aware of any examples of real-world evolution of both control and morphology for complex legged robots.

\section{Robot and evolutionary setup} \label{sec.implementation}
In this section we present the physical robot and its control system, the evolutionary setup, and the physical test setup we use in our experiments. 

\subsection{The robot}
A custom robotic platform (shown in Figure \ref{fig.robot}) was used for all experiments in this paper, and is currently under development at the University of Oslo.
Details on the platform itself can be found in our previous work~\cite{tonnesfn_IROS18}, and we have previously used it for evolving control with static morphology~\cite{tonnesfn_ices16}.
The top frame measures about 480mm by 300mm, connecting the four legs in a mammalian configuration.
All legs have the ability to change their length, with a minimum length of 550mm, and maximum length of 670mm.
The middle link, or femur, has a minimum length of 185mm and a maximum of 210mm, while the lowest link, tibia, has a minmum length of 255mm and a maximum of 350mm.
  
Each leg includes three Dynamixel MX-64AT servos, with onboard PID controllers to receive the angle commands over USB. 
These servos are powered at different voltages in the experiments, and their operating characteristics are shown in Table \ref{table.servoParams}.
Reducing the voltage from the optimal voltage at 14.8V to a reduced voltage of 12V limits both torque and control by about 20\%.

\begin{table}
  \centering
  \caption{Characteristics of the Dynamixel MX-64AT servos when powered at different voltages.}
  \label{table.servoParams}
  \begin{tabular}{  c  c  c }
    \hline
    \bfseries Parameter name & \bfseries 12V & \bfseries 14.8V \\
    \hline
    No load speed & 63rpm & 78rpm \\
    \hline
    Stall torque & 6.0Nm & 7.3Nm \\
    \hline
    Stall current draw & 4.1A & 5.2A \\
    \hline
    Stall power draw & 49.2W & 78.0W \\ 
    \hline
  \end{tabular} 
\end{table}
  
The reconfigurable legs use small DC motors connected to lead screws, with aluminium rails for mechanical strength.
An Arduino Mega with a custom shield is used for the control, and we achieve a sub-millimetre accuracy on the leg length.
The low speed of reconfiguration ($\approx$1mm/s) makes it ineffective to use these actively during the gait, so they are exclusively used for changing morphology, and are not seen by the controller.
  
An Xsens MTI-30 \textit{Attitude and Heading Reference System} (AHRS) is mounted close to the middle of the body to measure linear acceleration, rotational velocity and magnetic fields, giving data on absolute orientation at 100Hz. 
Reflective markers are mounted on the main body of the robot to allow motion capture equipment to record the position and orientation of the robot at 100Hz. 
The complete robot weighs 5.5kg, and operates tethered during all experiments.
  
\subsection{Control system}
We use a high-level inverse-kinematics based position controller for the legs of the robot.
The platform also supports a low-level controller, but this is only used in simulation experiments, due to the high number of evaluations needed before stable gaits are found.
A continuous, regular crawl gait \cite{GonzalezdeSantos2006} was chosen, where the body moves at a constant forward speed during the gait sequence, and lifts each leg separately to maximize stability. 
This setup allows gaits that are statically stable, although the low weight of the legs in relation to the body makes achieving faster gaits without introducing dynamic effects challenging.
The path for each individual leg end is defined by a Catmull-Rom spline.

The gait generator uses parameter ranges defined in Table \ref{table.params} and generates a number of control points for the spline, resulting in a continuous gait path for each leg\footnote{Details on control point generation can be found in the source code at \url{http://robotikk.net/project/dyret/}}.
Three parameters are used for manipulating the control points. The parameter step\_length controls the length of the ground contact line, while step\_height determines the height of the step. 
The step\_smoothing parameter regulates the angle of movement at the point where the leg hits the ground, by stretching out the front of the spline. 
This was added to allow for a reduction of the impact forces from each step, by making contact with the ground in a more horizontal direction.

\begin{table}
\centering
\caption{Gait parameters. These have been constrained (*) to limit the robot to a maximum speed of 10m/min.}
\label{table.params}
\begin{tabular}{  l  l  l }
  \hline
  \bfseries Category               & \bfseries Name             & \bfseries Values \\ \hline
  \multirow{2}{60pt}{Spline shape} & \textit{step\_length}      & [ 5mm, 300mm]*   \\ \cline{2-3}
                                   & \textit{step\_height}      & [25mm,  75mm]    \\ \cline{2-3}
                                   & \textit{step\_smoothing}   & [   0,  50mm]    \\ \cline{2-3}
  \hline
  \multirow{2}{60pt}{Gait timing}  & \textit{gait\_frequency}   & [0.2Hz,  2Hz]*   \\ \cline{2-3}
                                   & \textit{lift\_duration}    & [  5\%,  20\%]   \\
  
  \hline
  \multirow{2}{60pt}{Balancing}    & \textit{wag\_phase}        & [$-$0.2, 0.2]    \\ \cline{2-3}
                                   & \textit{wag\_x\_amp}       & [0, 50mm]        \\ \cline{2-3}
                                   & \textit{wag\_y\_amp}       & [0, 50mm]        \\ \cline{2-3}
  \hline
  \multirow{2}{60pt}{Morphology}   & \textit{femur\_length}     & [0, 25mm]        \\ \cline{2-3}
                                   & \textit{tibia\_length}     & [0, 95mm]        \\ \cline{2-3}           
  \hline
  \vspace{-3.5mm}
\end{tabular}
\end{table}

To increase the stability of the gait, a configurable balancing ``wag'' movement was added where the robot leans to the opposite side of the currently lifted leg. 
This ensures a higher margin of stability, and is required for a statically stable gait due to the relatively high mass of the legs compared to the body. 
Parameters for the phase and amplitude of the balancing wag can be changed individually for the lengthwise and sideways movement.

The maximum theoretical speed of the robot is given by the gait\_frequency and step\_length parameters; however, the actual speed of the robot also depends on its stability, and friction between its feet and the ground.
Setting a high gait\_frequency and low step\_length, and also a low gait\_frequency and high step\_length would result in valid gaits. If both parameters are set too high, however, the robot might end up damaging itself by trying to achieve a non-realistic forward speed. 
We therefore limit the product of step\_length and gait\_frequency to 10m/min.
The lift\_duration parameter decides how much of the gait period is used to lift the leg through the air, before beginning the next step.

The gait is made completely independent of the robot morphology by sending the goal position of the legs to an inverse kinematics function that reads the lengths of the legs at 10Hz.
No adaptation of any kind is done in the controller for the different morphologies, as we do not want to impose any limitations based on a priori knowledge of the design.
It might, for instance, be intuitive that an individual with longer legs might work better taking longer steps, but we do not want to add more dependencies between morphology and control than exists naturally in the system.
Minimizing the dependencies makes it easier to analyse the results, as there are fewer factors affecting the evolutionary search and its results.

The control system is implemented in C++ and uses the software framework \textit{Robot Operating System} (ROS) \cite{ros}. 
The leg end positions from the gait controller are sent through an inverse kinematics function to obtain the angles necessary to achieve the specified pose. 
The different functions of the robot controller are implemented as individual ROS nodes, and run on a computer connected to the robot by cable. 

\subsection{Evolutionary setup}

Mammal-inspired four-legged robots, as used in this work, are more prone to fall than spider- or lizard-inspired robots commonly used in evolutionary robotics.
Our robot's narrow stance, downward extending legs, and high centre of gravity, present much more danger of falling to the side than other bio-inspired designs. 
To be able to evolve fast gaits that are also robust on our platform, it is important to include stability as a fitness objective, in addition to speed.
These two goals are conflicting, as a robot standing still has great stability, while a fast robot necessarily has some movement that will be interpreted as instability.
We therefore chose the NSGA-II algorithm \cite{nsga} to identify a Pareto front of solutions; a number of gaits with different trade-offs between the two objectives.
The software running the evolutionary algorithm uses Sferes2 \cite{sferes2}, a C++ framework for evolutionary experiments. 

Parameters are represented as real numbers with the values shown in Table \ref{table.params}. 
Gaussian mutation is used on all genes with an initial sigma of 1/6, which decays per generation to enhance the exploration early in the search, but still allow exploitation in later generations.
These meta-parameters were tuned to perform well at the low number of evaluations used in our experiments.
Since both exploration and exploitation is covered by the mutation, we use no recombination.
The step\_length and gait\_frequency are further limited by a maximum theoretical speed of 10m/min.
If after mutation the gait surpasses this limit, mutation is done again until it is within the limits.
Three runs are done for each experiment, and they all contain 8 generations of 8 individuals each, for a total of 192 evaluations for each experiment.
When re-evaluating single individuals, they are evaluated 10 times each to get a satisfactory statistical distribution of their fitness in the real world. 
To avoid effects on the performance due to setup, we did our re-evaluations on a different day than the original evolutionary runs.
The evolutionary parameters are summed up in Table \ref{table.evoParams}.

\begin{table}[ht]
  \centering
  \caption{Parameters for the evolutionary experiments}
  \label{table.evoParams}
  \begin{tabular}{  l   l }
    \hline
    \bfseries Name & \bfseries Value \\
    \hline
    Algorithm & NSGA-II \\
    \hline
    Evaluation time & Maximum 60s \\
    \hline
    Parameters & Real: [0, 1] \\
    \hline
    Recombination & None \\
    \hline
    \multirow{5}{60pt}{Mutation}    & Type: Gaussian \\  \cline{2-2}
                                    & Probability: 1.0 \\ \cline{2-2}
                                    & Initial sigma: 1/6 \\ \cline{2-2}
                                    & Sigma decay per generation: 0.05\\ \cline{2-2}
                                    & Minimum sigma: 0.05\\
    \hline
    \multirow{2}{60pt}{Evaluations} & Population: 8 \\  \cline{2-2}
                                    & Generations: 8 \\ \cline{2-2}
                                    & Runs per experiment: 3\\ \cline{2-2}
                                    & Evaluations per re-evaluation: 10 \\
    \hline
    \vspace{1mm}
  \end{tabular} 
\end{table}

Two fitness functions are used in the experiments in this paper, speed and stability. 
The speed is calculated by using the duration of the gait and the Euclidean distance between the start and end position captured by the motion capture equipment, as seen in Equation \ref{eq.fitSpeed}, resulting in a measure of traversed meters per minute.  
We use a fitness function for stability based on the orientation and measured linear acceleration from the AHRS.
The full stability objective function, seen in Equation \ref{eq.fitStab}, is a weighted sum of the linear acceleration and orientation function, where \textit{acc} are samples from the accelerometer, \textit{ang} are samples from the orientation output of the AHRS, \textit{i} is the sample index, and \textit{j} is the axis of the sample. 
The accelerometer records data in the x, y and z-axes, while orientation is recorded in roll, pitch and yaw.
The scaling factor $\alpha$ was chosen to provide a balance between the two stability measurements by having acceleration and orientation affect the fitness value equally, and was in these experiments set to $0.02$.
The stability objective function is negated to allow for maximization of both objective functions, which means that a perfectly stable robot has a stability score of 0.
Samples in both functions are recorded at 100Hz.

\begin{equation}
  F_{speed} = \frac{\lVert P_{end} - P_{start} \rVert}{time_{end} - time_{start}}
  \label{eq.fitSpeed}
\end{equation}

\vspace{8mm}

\begin{equation*}
  G(A_{j}) = \sqrt{\frac{1}{n}\sum\limits_{i=1}^{n} (A_{j,i}^2-\overline{A_{j}}^2)}\\
\end{equation*}

\vspace{-1.5mm}

\begin{equation}
  F_{stability} = -\left(\alpha * \sum\limits^{axes}{G(Acc_{axis})} + \sum\limits^{axes}{G(Ang_{axis})} \right)
  \label{eq.fitStab}
\end{equation}

\subsection{Physical test setup and evaluations}
The goal of the physical test setup is to maximize the quality of measurements, while minimizing down time and requirements for human intervention. 
Motion capture equipment is used to provide a precise and accurate reading of position for estimation of speed. 
The duration of each gait test is chosen to provide a good balance between the number and accuracy of evaluations, given the time budget.
Each evaluation is obtained by walking one and a half meters forward, and then walking back to the start position using the same gait in reverse, before averaging the fitness values achieved in both directions. 
Each path is restricted by a timeout of 15 seconds, to limit the time spent on evaluating the slower individuals.
Evaluating a gait both directions help cancel out any asymmetric dynamics in the system that is caused by minor differences in the mechanics of the left and right side of the robot.

Both the robot and control system are designed to ensure repeatability for gaits by keeping the distance moved between each evaluation minimal.
This is achieved by having the robot sequentially lift and reposition the legs to the start pose of new gaits after each evaluation. 
Two walking sequences of 15 seconds, in addition to mechanical reconfiguration and repositioning of legs before and after the gait, results in a maximum of about 60 seconds used for each evaluation.
Some human intervention is required if the robot falls, or gets too close to the perimeter of the experiment area.
In practice, such intervention seems to be required every one to five minutes, depending on the objectives used and stage of evolution. 
If the robot falls or finishes evaluation without being parallel to the floor, the program pauses and waits for human intervention before continuing, to ensure only valid fitness scores are recorded.

\section{Experiments and results} \label{sec.experiments}  
Our main experiment is comprised of evolutionary multi-objective runs at the two different voltage levels. 
We compare the fitness from the two groups of runs, and examine the resulting individuals to identify signs of adaptation of control and morphology in the populations.
A selection of individuals from the optimal voltage runs is then re-evaluated 10 times each to gain a representative measurement of their fitness. 
This re-evaluation is done at both optimal and reduced voltage to determine how the change in supply voltage affects performance, and to shed light on the need for adaptation when subjected to this change.
In this section, we first present the results of the main experiment, before showing the results from the re-evaluation of individuals.

\subsection{Evolutionary runs}

\begin{figure}
  \includegraphics[width=0.46\textwidth]{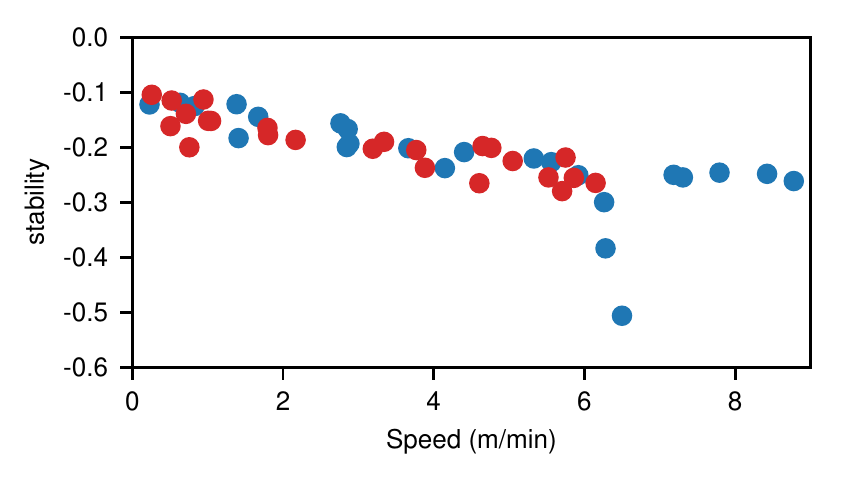}
  \caption{Comparison between fitness of the last generations evolved with optimal and reduced voltage. All individuals are optimized towards the top right, where an individual would have both high speed and stability.}
  \label{fig.finalpopcomp}
  \vspace{-2.5mm}
\end{figure} 

\begin{figure}
    \begin{subfigure}{0.5\textwidth}
        \includegraphics[width=8.5cm]{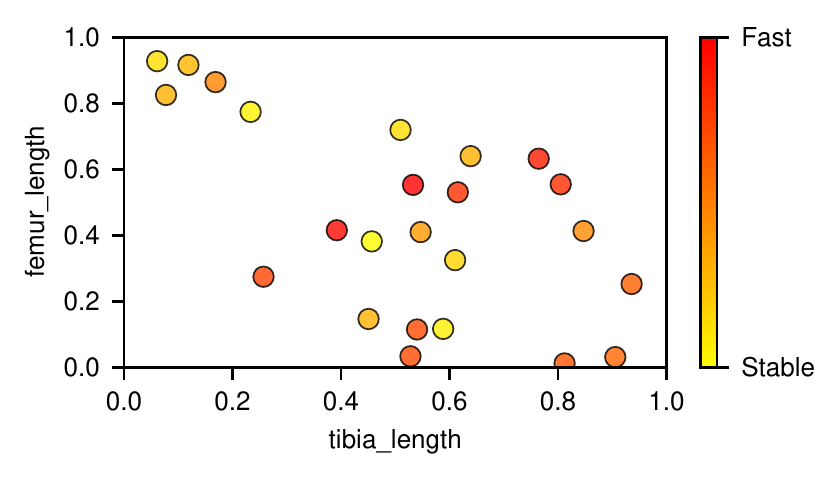}
        \caption{Optimal voltage leg lengths.}
        \label{fig.legLengths15}
    \end{subfigure} 
    
    \begin{subfigure}{0.5\textwidth}
        \centering
        \includegraphics[width=8.5cm]{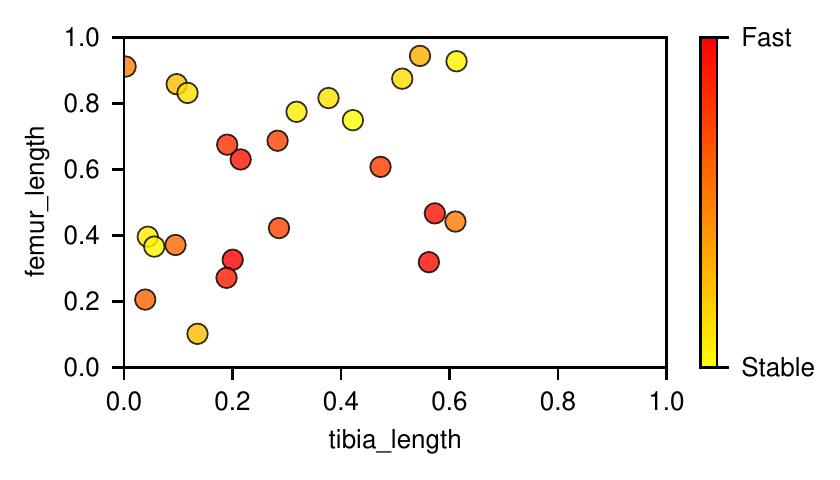}
        \caption{Reduced voltage leg lengths.}
        \label{fig.legLengths12}
    \end{subfigure}
    \caption{The length of the two reconfigurable leg segments for the last generations of the evolutionary runs. The colour indicates where in Fig. \ref{fig.finalpopcomp} the individual comes from, with the fastest robots in red, and the most stable robots in yellow.}
    \label{fig.legLengths}
\end{figure}

The last populations of all runs are shown in Figure \ref{fig.finalpopcomp}.
The optimal voltage runs achieve a higher speed of up to 9m/min, while the reduced voltage runs achieve speeds of just over 6m/min.
Even though only optimal voltage individuals achieve high speeds, the performance of both runs is comparable for small and moderate speeds.
The final populations for both groups have a reasonably linear shape broken only by a low-stability tail at around 6m/min in one of the optimal voltage runs.

Figure \ref{fig.legLengths} shows the morphologies that resulted from the runs with the two different voltages.
The colour of the individuals shows the difference in fitness of the individuals, showing the relationship between morphology and achieved speed and stability in the experiments.
For the optimal voltage individuals in Figure \ref{fig.legLengths15}, we see a smaller clustering of high femur length and low tibia length individuals, and a larger clustering of high tibia length with moderate to low femur length individuals.
They use a maximum of 79\% of the available reconfigurable leg length, while the mean individual uses 50\% of its available reconfigurable length.

For the reduced voltage runs in Figure \ref{fig.legLengths12}, we see that individuals use the whole range of reconfigurable femur length, but only up to about 60\% of the reconfigurable tibia length. 
Since the reconfigurable length of the tibia is much longer than for the femur, we only see up to 68\% of the total available reconfigurable leg length being used, with a mean of about 35\% for the reduced voltage.
We also see from the graphs in Figure \ref{fig.legLengths} that the performance of the individuals is not proportional to the total length of the robot, as several of the tallest robots only have moderate speeds, and a couple of the shorter individuals have some of the faster speeds.

\begin{figure*}[!ht]
  \centering
  \includegraphics[width=17.8cm]{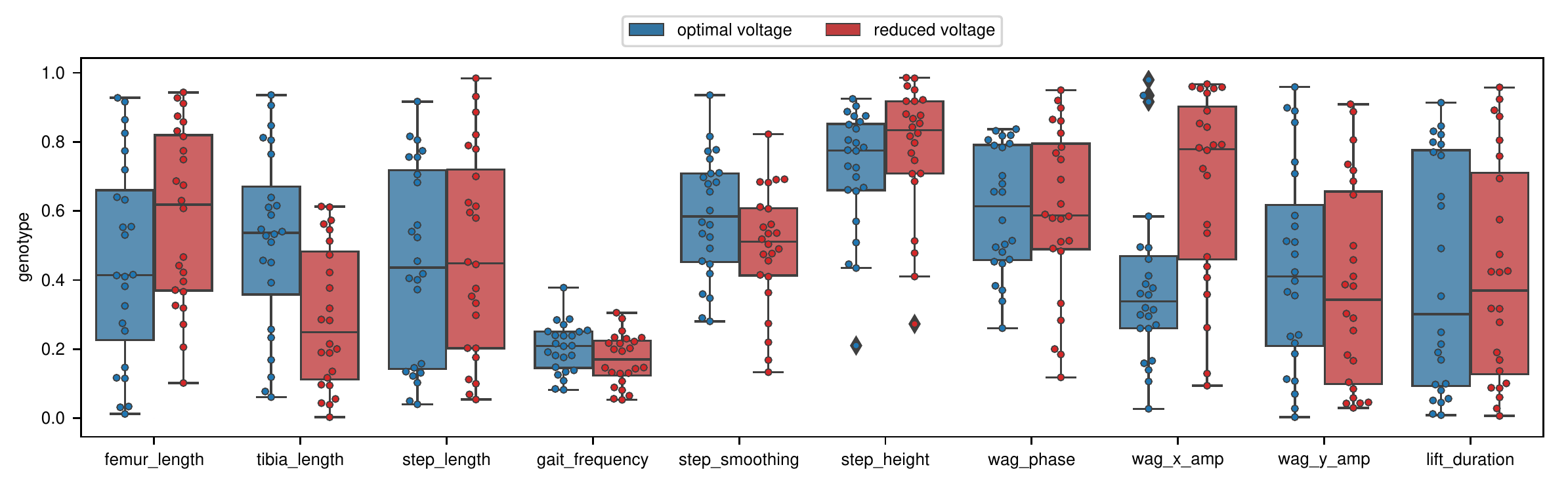}
  \caption{Genotype values and distributions for all individuals in the final generations resulting from the evolutionary runs.} 
  \label{fig.genotypecomp}
  \vspace{-2mm}
\end{figure*}

The boxplot in Figure \ref{fig.genotypecomp} reveals some of the differences in control and morphology parameters between the populations. 
There are clear differences in the tibia\_length and wag\_x\_amp parameters, and moderate differences in femur\_length, step\_smoothing, and step\_height. 
A detailed study of how each of the ten parameters is affected by the hardware limitations is out of the scope of this paper, so these differences are not investigated further individually.

However, we wish to analyse them on a group basis, in order to study the differences in morphology and control between the optimal and reduced voltage runs. 
To achieve this, linear discriminant analysis (LDA) was applied separately to the morphology and control parameters to give a one-dimensional representation of each group. 
This was followed by a Mann-Whitney U test to establish significance. 
The Mann-Whitney U test indicated a significant difference in the one-dimensional reduction of the two parameters for morphology, femur\_length and tibia\_length, due to the change in voltage ($U=138$, $p<0.01$), with Cliff's delta effect size of $-0.52$.
The same analysis on the eight control parameters reduced to one, also indicated significant differences ($U=92$, $p<0.01$), with a Cliff's delta effect size of $-0.68$.

\subsection{Re-evaluation of individuals}
Since we are using a high-level controller, it can be hard to directly predict how a change to a robots internal or external environment affects it, and we need to verify if adaptation is actually necessary when changing the voltage, or if the controller is able to handle both scenarios.
For this, we chose five individuals with different fitnesses from the optimal voltage runs. 
These were then evaluated ten times at their original voltage, before being tested again at the reduced voltage.
Re-testing under the original conditions is important to give an accurate comparison, as the noise in hardware measurements means that the single evaluation during evolution might not be representative of its true performance.
  
\begin{figure}
  \includegraphics[width=8.5cm]{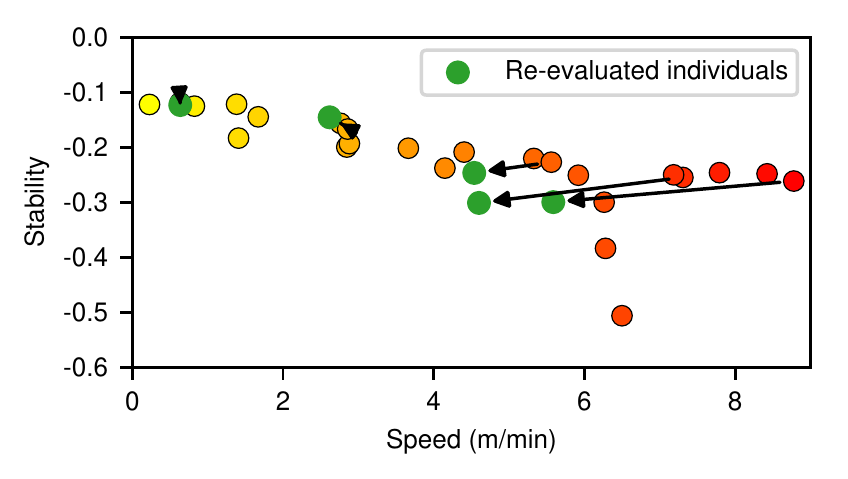}
  \caption{Fitness of individuals compared to their re-evaluations at reduced voltage. Green circles show the fitness at reduced voltage, and the black arrows show the change in fitness for these individuals.}
  \label{fig.reevals}
\end{figure}

The results are summarized in Table \ref{table.reevals}.
We can observe that for the two slowest individuals, the stability actually increases, while the stability decreases by 13\% to 17\% for the others. 
All mean speeds decrease, with the biggest reduction at 38\%.
All changes, except in the speed of the slowest individual, were shown to be statistically significant ($p<0.01$) using the Mann-Whitney U test with Holm-correction of the $p$-values. 
Figure \ref{fig.reevals} shows the change in fitness from original to reduced voltage, where green circles denote re-evaluated individuals at the lower voltage. 
This figure reveals the large drop in speed for fast individuals particularly clearly.

\begin{table}[ht]
  \vspace{5mm}
  \caption{Means and standard deviations of results from the re-evaluation of selected individuals. The original fitness from the evolutionary run is included, in addition to re-evaluated fitness at both optimal and reduced voltage. (* Statistically significant difference)}
  \label{table.reevals}
  \centering
  \resizebox{.48\textwidth}{!}{%
  \begin{tabular}{ c  c  c  c  @{\hskip 5mm}  c  c  c  c  }
    \hline
      \multicolumn{4}{c }{\bfseries Speed} & \multicolumn{4}{c }{\bfseries Stability} \\
      \textit{evo} & \textit{optimal} & \textit{reduced} & \textit{change} & \textit{evo} & \textit{optimal} & \textit{reduced} & \textit{change} \\
      \hline
    0.63 & $0.65\pm0.02$  & $0.64\pm0.02$ & \color{red_plot}{-1.9\%}   & -0.11 & $-0.13\pm0.00$ & $-0.12\pm0.00$ &  \color{green_plot}{+8.2\%*} \\ 
      \hline
    2.76 & $2.80\pm0.03$ & $2.62\pm0.02$ & \color{red_plot}{-6.5\%*}  & -0.16 & $-0.16\pm0.00$ & $-0.15\pm0.00$ &  \color{green_plot}{+6.4\%*} \\ 
      \hline 
    5.56 & $5.54\pm0.06$ & $4.54\pm0.24$ & \color{red_plot}{-18.0\%*} & -0.23 & $-0.22\pm0.01$ & $-0.25\pm0.01$ & \color{red_plot}{-12.8\%*} \\ 
      \hline
    7.31 & $7.15\pm0.13$ & $4.60\pm0.26$ & \color{red_plot}{-35.7\%*} & -0.26 & $-0.26\pm0.01$ & $-0.30\pm0.01$ & \color{red_plot}{-14.9\%*} \\ 
      \hline
      8.78 & $8.96\pm0.13$ & $5.59\pm0.17$ & \color{red_plot}{-37.6\%*} & -0.26 & $-0.26\pm0.02$ & $-0.30\pm0.01$ & \color{red_plot}{-16.7\%*} \\ 
      \hline
  \end{tabular}}
  \vspace{1mm}
\end{table} 

\FloatBarrier

\section{Discussion} \label{sec.discussion}
The decrease in performance seen in Figure \ref{fig.reevals} shows that lowering the supply voltage of the system affects the robot's gait. 
Reducing both torque and speed of the robot joints yielded a speed reduction of up to 38\% and a stability decrease of up to 17\%.
This large discrepancy shows the need for adaptation to keep performing well in dynamic environments with changing hardware conditions.
There is a large number of factors that can affect the performance of a robot, and it is likely that many robots, especially if working in complex environments or alongside other agents, might see similar, or even larger, differences in performance than we saw here.
A robot can adapt to some of these factors using the evolutionary techniques shown in this paper, but they have not been tuned to respond quickly to abrupt changes, and are only meant as an off-line adaptation to new hardware limitations or environments.

We see from the difference in Figures \ref{fig.legLengths15} and \ref{fig.legLengths12} that the lower powered individuals are not able to exploit the full available length of the legs.
This is supported by the fact that the mean reconfigurable leg length is 50\% for optimal voltage runs, and only 35\% for reduced voltage runs. 
Lower leg lengths can be seen as a gearing of the motors, as shorter legs trade speed for torque, and a reduction in leg lengths can therefore be seen as a response to the reduced torque.
An interesting detail shown in Figure \ref{fig.genotypecomp} is that even though results from the optimal voltage runs have a higher mean leg length, the femur\_length is generally highest in the reduced voltage runs.
Even though the interaction of these parameters under evolutionary optimisation is very complex, and might require more experiments to be understood fully, we still see a significant change in both morphology and control, which shows that the evolutionary search is able to adapt to the new hardware conditions by utilizing both.

The number of evaluations performed in this real-world study is limited compared to simulated ER research.
Early experiments showed little to no improvement in fitness past the sixth generation, so we chose to do eight generations for a high probability of the search to converge. 
We also saw that the resulting populations contained a good number of individuals with different trade-offs between the different objectives, indicating that we had sufficient population size. 
Considering Figure \ref{fig.finalpopcomp}, we see that there isn't a big difference in performance of the final populations between the runs, and we consider it unlikely that more runs would change the results considerably. 
Figure \ref{fig.genotypecomp} shows a large diversity in final populations for the two groups of runs.
We would expect to see much smaller variations for a converged evolutionary search with one objective, but that is not the case when doing multi-objective evolution using NSGA-II.
This algorithm has a mechanism for maximising the fitness diversity in each front of the population, and since our two fitness objectives are conflicting, we end up with a range of different individuals with different trade-offs between these two objectives, which necessarily results in higher diversity in the populations as well.

Evolving robots in real-world environments is often challenging due to noise in measurements.
The standard deviations in Table \ref{table.reevals} showed only small variations of performance in our experiments, even when the re-evaluations was done on another day.
These results confirm that we limited noise and uncertainty in our measurements to an acceptable level.

Figure \ref{fig.reevals} shows that only the faster individuals suffer significant losses in fitness and that more stable individuals are robust to the reduced supply voltage.
Visual observation of the evaluations suggested that the reduction in performance is most likely caused by the lower stability.
The theoretical speed of the gait is given by the high-level controller and the gait\_frequency and step\_length parameters, but unstable gaits stumble or miss steps, leading to lower distances covered in the same time.
This indicates that if we are to deploy this robot in new conditions, it might be wise to select more stable gaits, as they are most likely more robust to unknown environments.

\section{Conclusions and future work} \label{sec.conclusion}

In this paper we investigated the effects of lowering servo torque and speed on evolved robots, and to what degree the robot through evolutionary techniques was able to adapt to this change. 
This large reduction in performance from lowering the voltage shows the need for adaptation to keep performing well in dynamic environments with changing hardware conditions.
We showed that the evolutionary search was able to achieve comparable results to the original run at low and moderate speeds by changing both the control and morphology of the robot.
We also demonstrated the feasibility of doing multi-objective exploratory morphology and control evolution entirely in hardware on our new platform.

An avenue for future expansion of this work would be to further investigate the actual contribution from using evolutionary algorithms over random search, and investigate other techniques from machine learning to implement on-line optimization as well. 
The adaptation to lower servo torque and speed in this paper has been done off-line, and we expect that doing this adaptation on-line instead would pose additional challenges with interesting solutions and results.
Adding closed-loop control, opening up more parameters in the control system, or having separate parameters for each leg would give the system more possibilities for adapting, though getting feasible gaits in the start of the search with a mammal-inspired configuration can be very challenging.
Current methods for generating behavioural repertoires could benefit from dynamic morphologies.
It may also be possible to reduce the need for human intervention, allowing experiments in even more complex environments, encouraging investigations into embodied cognition and the interactions between robot body, mind, and environment.

We showed that our evolutionary system is able to adapt both control and morphology to new hardware limitations, but also that it is possible to do multi-objective exploratory morphology and control evolution in relatively few evaluations entirely in hardware, hopefully inspiring more researchers to take the leap into real world evolutionary experiments.

\begin{acks}
This work is partially supported by The Research Council of Norway as a part of the Engineering Predictability with Embodied Cognition (EPEC) project, under grant agreement 240862.
\end{acks}

\bibliographystyle{ACM-Reference-Format}
\bibliography{bibliography} 

\end{document}